\pdfoutput=1

\documentclass[11pt]{article}

\usepackage[final]{acl}

\usepackage{times}
\usepackage{latexsym}

\usepackage[T1]{fontenc}

\usepackage[utf8]{inputenc}
\usepackage{booktabs}
\usepackage{microtype}
\usepackage{overpic}
\usepackage{inconsolata}

\usepackage{graphicx}
\usepackage{tabularx}
\usepackage{adjustbox}
\usepackage{multirow}
\usepackage{makecell}

\newcommand{\struct}{\textit{MoDULA}}
\newcommand{\structone}{\textit{MoDULA-Flan}}
\newcommand{\structtwo}{\textit{MoDULA-Res}}
\newcommand{\molora}{MoLoRA}

\author{
    \textbf{Yufei Ma\textsuperscript{1,\thanks{Equal Contribution.}}}\ \ \ \ \ 
    \textbf{Zihan Liang\textsuperscript{3,\footnotemark[1]}}\ \ \ \ \ 
    \textbf{Huangyu Dai\textsuperscript{3,\footnotemark[1]}}\ \ \ \ \ 
    \textbf{Ben Chen\textsuperscript{3,\thanks{Corresponding Author.}}}
    \\
    \textbf{Dehong Gao\textsuperscript{1,4,\footnotemark[2]}}\ \ \ \ \ 
    \textbf{Zhuoran Ran\textsuperscript{2,3}}\ \ \ \ \ 
    \textbf{Zihan Wang\textsuperscript{3}}\ \ \ \ \ 
    \textbf{Linbo Jin\textsuperscript{3}} 
    \\
    \textbf{Wen Jiang\textsuperscript{3}}\ \ \ \ \ 
    \textbf{Guannan Zhang\textsuperscript{3}}\ \ \ \ \ 
    \textbf{Xiaoyan Cai\textsuperscript{2}}\ \ \ \ \   
    \textbf{Libin Yang\textsuperscript{1}}  
    \\
    \textsuperscript{1}{Northwestern Polytechnical University, School of Cybersecurity, Xi'an, China}\\
    \textsuperscript{2}{Northwestern Polytechnical University, School of Automation, Xi'an, China} \\
    \textsuperscript{3}{Alibaba Group, Hangzhou, China} \\
    \textsuperscript{4}{Binjiang Institute of Artificial Intelligence, ZJUT, Hangzhou, China}
    \\
}

\title{MoDULA: \underline{M}ixture \underline{o}f \underline{D}omain-Specific and \underline{U}niversal \underline{L}oR\underline{A} for Multi-Task Learning}

\begin{document}

\maketitle

\begin{abstract}
The growing demand for larger-scale models in the development of \textbf{L}arge \textbf{L}anguage \textbf{M}odels (LLMs) poses challenges for efficient training within limited computational resources. 
Traditional fine-tuning methods often exhibit instability in multi-task learning and rely heavily on extensive training resources.
Here, we propose \struct~(\textbf{M}ixture \textbf{o}f \textbf{D}omain-Specific and \textbf{U}niversal \textbf{L}oR\textbf{A}), a novel \textbf{P}arameter \textbf{E}fficient \textbf{F}ine-\textbf{T}uning (PEFT) \textbf{M}ixture-\textbf{o}f-\textbf{E}xpert (MoE) paradigm for improved fine-tuning and parameter efficiency in multi-task learning. 
The paradigm effectively improves the multi-task capability of the model by training universal experts, domain-specific experts, and routers separately. \structtwo~is a new method within the \struct~paradigm, which maintains the model's general capability by connecting universal and task-specific experts through residual connections.
The experimental results demonstrate that the overall performance of the \structone~and \structtwo~methods surpasses that of existing fine-tuning methods on various LLMs. 
Notably, \structtwo~achieves more significant performance improvements in multiple tasks while reducing training costs by over 80\% without losing general capability. 
Moreover, \struct~displays flexible pluggability, allowing for the efficient addition of new tasks without retraining existing experts from scratch. This progressive training paradigm circumvents data balancing issues, enhancing training efficiency and model stability. Overall, \struct~provides a scalable, cost-effective solution for fine-tuning LLMs with enhanced parameter efficiency and generalization capability.
\end{abstract}

\begin{figure*}[ht]
  \centering
  \includegraphics[width=\linewidth]{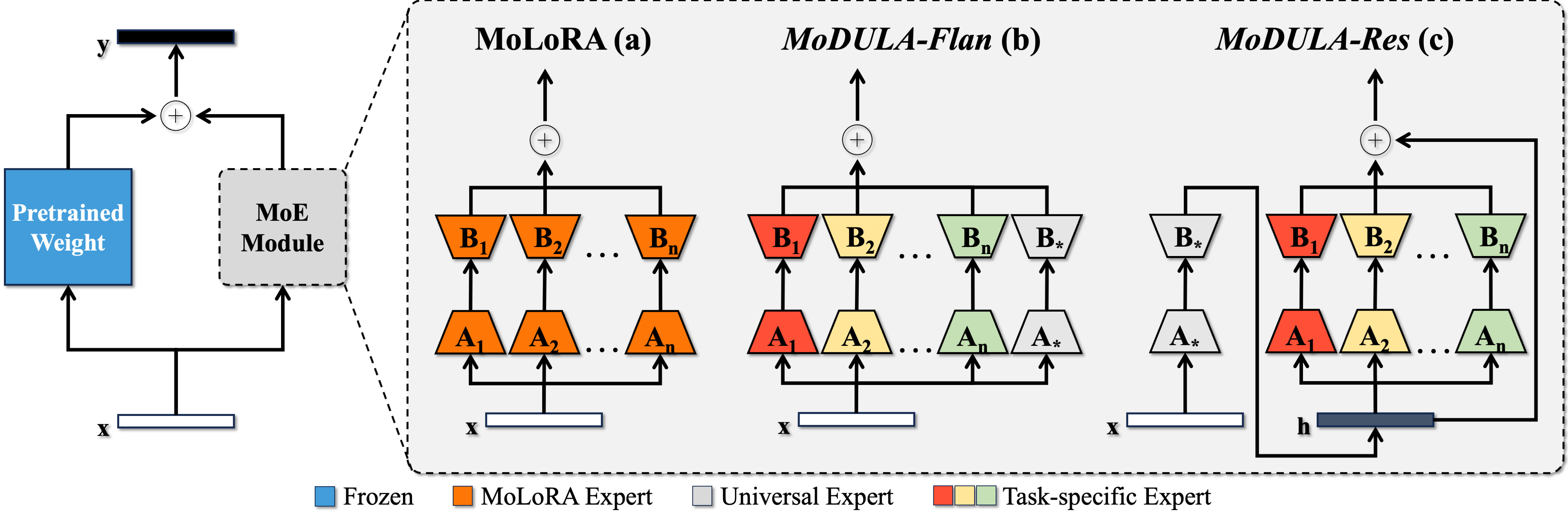}
  \caption{Illustrations of \molora(a), \structone(b), and \structtwo(c) with router omitted.}
  \label{fig:comparison}
\end{figure*}

\section{Introduction}
\par
Recent advancements in open-source Large Language Models (LLMs), such as LLaMA~\citep{touvron2023llama}, Qwen~\citep{bai2023qwen}, and Yi~\citep{ai2024yi}, have achieved notable successes in natural language processing. 
However, the increasing complexity and growing size of these models make efficient training within limited computational resources challenging. 
Researchers tried to address this with Parameter Efficient Fine-Tuning (PEFT), such as LoRA~\citep{hu2021lora}, Prefix Tuning~\citep{liu2023ptuning}, and $(IA)^3$~\citep{liu2022ia3}. 
LoRA has gained prominence for its high performance using low-rank matrices, but it often encounters instability when trained on large, mixed datasets. 
To mitigate this issue, \molora~\citep{zadouri2024molora} has been introduced by extending LoRA and integrating the Mixture-of-Expert (MoE) architecture as shown in Figure~\ref{fig:comparison}(a). This approach trains multiple LoRA-adapters concurrently, each serving as an expert, to enhance the base LLMs' generalization ability across diverse tasks. 
The integration of MoE into LoRA aims to improve training efficiency and stability, facilitating more effective fine-tuning of large-scale language models for a wide range of natural language processing applications.

\par
{
Despite its advantages, \molora~has some limitations. 
One limitation is \textbf{the absence of domain-specific LoRA adapters}, as the same experts are employed universally across all tasks.
This uniformity may limit the performance ceiling, especially for significantly distinct tasks like math and code, where the inclusion of domain-specific experts could potentially enhance performance~\citep{zeng2021pangualpha}.
Another challenge is \textbf{the limited pluggability} of \molora; adding new task capabilities necessitates retraining all parameters from all experts, which can be inefficient and time-consuming.
}

\par
To address the challenges, we propose a three-stage training paradigm called \struct, where different domain-specific experts can be trained separately. 
Moreover, we introduce a more advanced method \structtwo~(\textbf{M}ixture \textbf{o}f \textbf{D}omain-Specific and \textbf{U}niversal \textbf{L}oR\textbf{A} with \textbf{Res}idual Connection), which incorporates a residual structure to make the training more stable, as seen in Figure~\ref{fig:comparison}(c).
Unlike \molora, which employs multiple identical LoRA adapters as experts, our paradigm incorporates a universal expert alongside multiple domain-specific experts. 
The universal expert learns task-agnostic representations, while each domain-specific expert operates as a bias adapter, focusing on domain-specific knowledge.
Intuitively, arranging these adapters in parallel and allocating weights to each adapter via a router constitutes the \structone~(\textbf{M}ixture \textbf{o}f \textbf{D}omain-Specific and \textbf{U}niversal \textbf{L}oR\textbf{A} with \textbf{Flan} Routing) method as seen in Figure~\ref{fig:comparison}(b).
However, this method may potentially compromise universal capabilities. To address this, \structtwo~ introduces a refined method that enables domain-specific experts to receive input from the output of the universal expert. 
This design ensures a coherent flow of information and facilitates the optimal integration of both universal and domain-specific expert functionalities through a residual connection. 
By dynamically adjusting the contributions of domain-specific experts, \structtwo~adapts to individual tasks while preserving broad generalization capabilities. This flexibility allows the model to leverage its general competencies for task understanding and summarization when encountering new tasks, thereby achieving a more balanced and effective adaptation in multi-task scenarios.

\par
{
During model training, our \struct~employs a three-stage optimization process, with detailed illustrations displayed in Figure~\ref{fig:training-stages}:
1) Initially, only the universal expert is trained to adapt to general tasks quickly;
2) Subsequently, each domain-specific expert is trained individually, focusing on its corresponding task;
3) Finally, the parameters of all experts are frozen, and only router is trained to learn the optimal combination strategy for different tasks.
This progressive training paradigm allows our methods to avoid retraining from scratch, distinguishing it from \molora, which trains only a new expert for a new specific task and retraining the router. 
This paradigm significantly reduces computational costs, mitigates data balancing challenges, and enhances the model's pluggability.
}

\par
{
To evaluate the effectiveness of our proposed methods, we conduct extensive experiments on a diverse set of open-source LLMs, including LLaMA-2~\citep{touvron2023llama2}, Qwen~\citep{bai2023qwen}, and Yi~\citep{ai2024yi}, across various tasks.}
{
The results consistently demonstrate that \struct~exhibits a significant performance,  achieving 4.5\% improvements compared to \molora. By introducing residual connections, \structtwo~achieves even greater improvements without compromising the general capabilities. Additionally, our approach showcases superior adaptability to new tasks, outperforming \molora~in finance and e-commerce domain with less training data and parameters, highlighting the enhanced task pluggability of our approach, making it an efficient and general solution for multi-task learning in LLMs.
}

\section{Related Works}
\subsection{Large Language Model}
{Recently, the field of natural language processing has witnessed a paradigm shift with the advent of LLMs~\citep{anil2023palm,falcon-180b,xu2023baize,scao2022bloom,brown2020gpt,openai2023gpt4,zhang2023internlmxcomposer,du2022glm}. 
These state-of-the-art models have departed from traditional approaches that relied on convolutional or recurrent architectures for feature extraction, instead embracing novel techniques such as BERT~\citep{devlin-etal-2019-bert}, which leverages the power of Transformers trained on extensive datasets, yielding bidirectional encoder representations. 
Similarly, Generative Pretrained Transformer (GPT)~\citep{brown2020gpt} employs decoder layers from Transformer architecture~\citep{vaswani2017attention} as feature extractors and utilizes autoregressive training on vast texts.
}

{
Guided by the principles of scaling laws~\citep{kaplan2020scaling}, the development of LLMs has led to the emergence of colossal models boasting over 100 billion parameters, with prominent examples including GPT-4~\citep{openai2023gpt4} and Gemini~\citep{geminiteam2023gemini}. 
Interestingly, open-source models such as OPT~\citep{zhang2022opt}, Falcon~\citep{falcon-180b}, and Gemma~\citep{gemmateam2024gemma} have demonstrated competitive performance compared to their closed-source counterparts, despite possessing a more modest parameter count.
The training process of LLMs typically involves leveraging immense amounts of textual data to enable the prediction of subsequent tokens, empowering these models to generate coherent and comprehensible responses to a wide range of prompts. 
This training method has proven to be highly effective in capturing the intricacies of language and paved the way for LLMs to achieve SOTA performance across various NLP tasks.
}

\subsection{MoE for PEFT}
\par
Our research closely aligns with the work done by \molora~\citep{zadouri2024molora}, LoraHub~\citep{huang2023lorahub}, MoELoRA~\citep{liu2024moelora_sigir}, SiRA~\citep{zhu2023sira}, and C-Poly~\citep{wang2023customizable}, which explore the intersection of PEFT and MoE.
\molora~employs a full soft MoE on top of LoRA, utilizing a learned gating mechanism to average all experts, and trains the experts in a single stage.
LoraHub investigates LoRA composability for cross-task generalization and introduces a simple framework for the purposive assembly of LoRA modules trained on diverse given tasks, aiming to achieve adaptable performance on unseen tasks. 
It can fluidly combine multiple LoRA modules with just a few examples from a new task, without requiring additional model parameters or human expertise.
MoELoRA devises multiple experts as the trainable parameters and proposes a task-motivated gate function for all MOELoRA layers to regulate the contributions of each expert and generate distinct parameters for various tasks.
SiRA proposes a sparse mixture of low rank adaption that enforces the top k experts' routing with a capacity limit. 
It uses expert dropout to reduce over-fitting.
C-Poly combines task-common skills and task-specific skills and jointly learns a skill assignment matrix.
\par
While these methods have significantly contributed to the field, they face particular challenges and limitations.
Training experts on mixed datasets as in \molora~may lead to performance degradation due to data inconsistency and interference~\citep{dong2024abilities}.
LoraHub relies on few-shot examples in inference stage, and MoELoRA requires task-id to determine which experts should be activated, which weaken the flexibility of both methods.
Sparse routing, as used by SiRA, requires careful tuning of the top-k and capacity hyperparameters for each dataset.
C-Poly's joint learning of task-common and task-specific skills can make balancing general and specialized abilities difficult.
Additionally, incorporating new experts or skills in these methods may require retraining or modifying existing components, potentially impacting system stability and training complexity.
Training new experts often demands substantial data, resulting in high training costs and sub-optimal performance in specific domains.
Maintaining optimal performance on domain-specific benchmarks after adding new capabilities can be challenging, and newly added modules may not consistently achieve top performance in their respective benchmarks.
These factors can affect the adaptability and efficiency of \molora, SiRA, and C-Poly in meeting expanding task demands.
\par
In contrast, \struct~method trains universal and domain-specific experts separately, mitigating performance degradation from mixed datasets. 
Designed with "pluggability" in mind, the \struct~method allows new experts to be added without changing existing ones, ensuring system stability and low training costs.
After adding a new expert, only the router requires retraining to maintain near-optimal performance. 
This staged training balances general and domain-specific capabilities, making our method adaptable and efficient for growing task requirements.

\begin{figure*}[htbp]
  \centering
  \includegraphics[width=\linewidth]{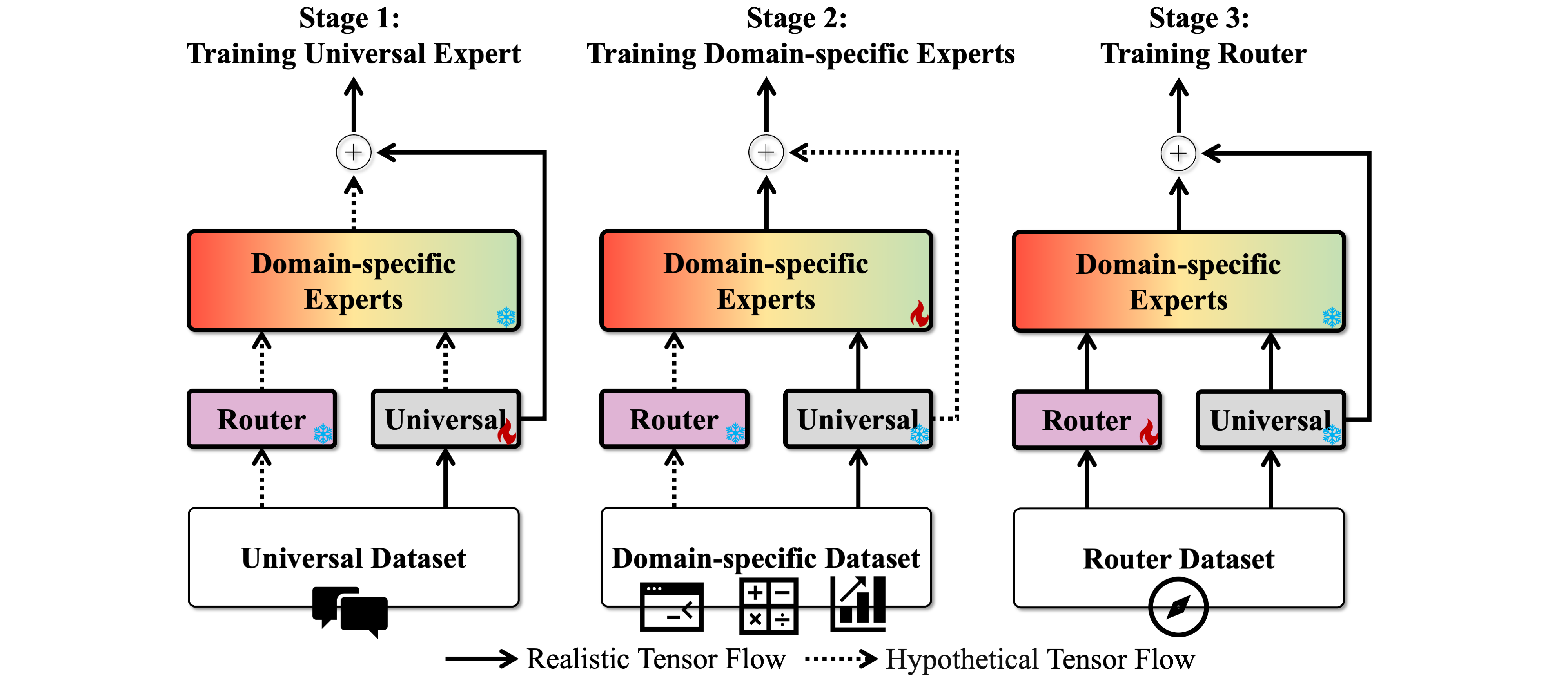}
  \caption{Illustrations of the three-stage training paradigm for \structtwo.}
  \label{fig:training-stages}
\end{figure*}

\begin{table*}[ht]
\centering
\scriptsize 
\renewcommand{\arraystretch}{1} 
\begin{adjustbox}{max width=\textwidth}
\begin{tabular*}{\textwidth}{@{\extracolsep{\fill}}cccccccccc}
    \toprule[1.5pt]
    \textbf{Base Model} & \textbf{Method} & \textbf{Avg.}  & \textbf{GSM8K} & \textbf{Arithmetic} & \textbf{MathQA} & \textbf{HumanEval} & \textbf{MBPP} & \textbf{Medical} & \textbf{MedQA}\\
    \midrule[1pt]
    \multirow{6}{*}{Qwen-7B} & Not fine-tuned & 44.65 & 46.63 & 56.65 & 35.48 & 21.95 & 32.00 & 76.00 & 43.83 \\
    ~ & LoRA & 25.93 & 7.21 &	49.61 &	26.40 &	9.15 &17.20 & 42.80 & 29.14 \\
    ~ & LoraHub & 49.37 & 44.81 & 86.33 & 37.09 & 22.40 & 29.60 & 81.00 & 44.38 \\
    ~ & \molora~& 48.94 &	48.21 &	78.49 &	37.42 &	23.78 &	32.78 &	79.20 &	42.73  \\
    \cmidrule(lr){2-10}
    ~ & \structone~& 50.32 &	\textbf{48.67} &	87.06 &	36.98 &	23.17 &	\textbf{33.60} &	78.40 &	44.38 \\
    ~ & \structtwo~& \textbf{51.36} &	46.63 &	\textbf{90.37} &	\textbf{37.98} &	\textbf{25.00} &	33.00 &	\textbf{82.00} &	\textbf{44.55} \\
    \midrule[1pt]
    
    \multirow{6}{*}{LLaMA-2-7B} & Not fine-tuned & 27.45 & 	13.72 & 	6.89 & 	29.41 & 	14.63 	& 18.00 & 	77.60 & 	31.89 \\
    ~ & LoRA & 15.40 &	1.29 &	2.69 &	22.48 &	0.00 &	0.00 &	53.40 &	27.97  \\
    ~ & LoraHub & 38.69 & 22.03 & 63.47 & 31.17 & 13.80 & 24.00 & 83.60 & 32.79 \\
    ~ & \molora~& 38.53 &	\textbf{23.12} &	60.87 &	30.48 &	15.24 &	21.40 &	83.60 &	\textbf{35.03} \\
    \cmidrule(lr){2-10}
    ~ & \structone~& 38.67 &	20.39 &	61.40 &	31.35 &	15.24 &	\textbf{24.40} &	84.20 &	33.69 \\
    ~ & \structtwo & \textbf{39.62} &	22.37 &	\textbf{70.66} &	\textbf{31.73} &	\textbf{15.24} &	22.80 &	\textbf{85.20} &	29.31 \\
    \midrule[1pt]
    
    \multirow{6}{*}{Yi-6B} & Not fine-tuned & 38.04 &	33.81 &	39.92 &	35.41 &	14.63 &	23.00 &	70.00 &	49.49 \\
    ~ & LoRA & 16.07 &	2.51 &	0.88 &	20.41 &	0.00 &	0.00 &	61.20 &	27.49 \\
    ~ & LoraHub & 46.77 & \textbf{35.97} & 82.03 & 35.50 & 14.24 & \textbf{24.80} & 84.60 & 50.28 \\
    ~ & \molora~& 41.49 & 34.87 & 46.50 & 34.50 & 16.46 & 23.20 & 82.80 & \textbf{52.08} \\
    \cmidrule(lr){2-10}
    ~ & \structone~& 45.09 & 35.25 & 73.85 & 35.88 & 12.20 & 23.00 & 84.80 & 50.66 \\
    ~ & \structtwo & \textbf{48.61} & 34.50 & \textbf{92.72} & \textbf{36.29} & \textbf{16.46} & 24.40 & \textbf{85.80} & 50.12 \\
    \midrule[1pt]

    \multirow{6}{*}{Qwen-14B} & Not fine-tuned & 54.55 & 61.87 & 69.32 & 44.42 & 24.39 & 43.80 & 85.60 &	52.47 \\
    ~ & LoRA & 55.58 &	56.86 &	\textbf{92.58} & 39.23 & \textbf{26.83} & 37.60 & 82.80 &	53.18 \\
    ~ & LoraHub & 57.47 & 66.74 & 88.91 & 43.91 & 24.32 & 38.10 & 86.20 & \textbf{54.12} \\
    ~ & \molora~& 56.79 &	63.38 &	83.56 &	44.48 &	26.22 &	41.40 &	85.80 &	52.71 \\
    \cmidrule(lr){2-10}
    ~ & \structone~& 56.95 &	63.53 &	83.19 &	\textbf{45.25} &	25.61 &	42.40 &	85.60 &	53.10 \\
    ~ & \structtwo & \textbf{58.42} &	\textbf{67.78} &	91.45 &	45.13 &	18.90 & \textbf{44.80} &	\textbf{88.00} &	52.87 \\
    \midrule[1pt]

    \multirow{6}{*}{LLaMA-2-13B} & Not fine-tuned & 41.71 &	23.28 &	80.28 &	32.53 &	15.24 &	27.20 &	70.60 &	\textbf{42.81} \\
    ~ & LoRA & 16.33 &	1.18 &	4.28 &	25.27 &	0.00 &	0.00 &	55.00 &	28.59 \\
    ~ & LoraHub & 44.01 & 34.21 & 72.15 & \textbf{36.17} & 14.23 & 26.20 & 84.20 & 40.92\\
    ~ & \molora~& 45.62 &	33.51 &	74.57 &	34.21 &	19.51 &	30.40 &	85.80 &	41.32 \\
    \cmidrule(lr){2-10}
    ~ & \structone~& 44.70 &	35.48 &	67.31 &	34.53 &	20.73 &	28.60 &	83.80 &	42.46 \\
    ~ & \structtwo & \textbf{47.93} &	\textbf{36.47} &	\textbf{84.26} &	35.18 &	\textbf{20.73} &	\textbf{31.20} &	\textbf{86.40} & 41.24  \\
    \midrule[1pt]

    \multirow{6}{*}{Yi-9B} & Not fine-tuned & 56.45 &	51.33 &	93.27 &	39.97 &	25.61 &	49.20 &	82.60 & 53.18 \\
    ~ & LoRA & 16.23 &	0.69 &	0.82 &	22.95 &	0.00 &	0.00 &	61.40 &	27.73 \\
    ~ & LoraHub & 58.54 & 54.13 & 89.47 & \textbf{42.21} & 33.13 & 53.10 & 85.20 & 52.56\\
    ~ & \molora~& 56.97 & 57.99 & 68.89 & 41.86 & 32.32 & 54.20 & 86.80 & \textbf{56.72} \\
    \cmidrule(lr){2-10}
    ~ & \structone~& 60.54 &	\textbf{60.04} &	96.36 & 41.47 &	29.88 &	\textbf{54.80} &	86.80 &	54.43 \\
    ~ & \structtwo & \textbf{60.55} &	59.06 &	\textbf{96.86} 	& 41.51 &	\textbf{34.15} &	51.20 &	\textbf{87.20} &	53.86 \\
    \bottomrule[1.5pt]
\end{tabular*}
\end{adjustbox}
\caption{Main experimental results of baseline methods, \structone, and \structtwo~on domain-specific benchmarks.}
\label{tab:main-table-domain-specific}
\end{table*}

\section{Method}
\par
In this section, we present \struct~for LLM fine-tuning. 
Within this paradigm, we propose two methods: \structone~and \structtwo. 
\structone~consists of a universal expert and an array of domain-specific experts, while \structtwo~further incorporates residual connections between the universal and domain-specific experts to enhance performance and stability.
Figure~\ref{fig:comparison} illustrates the differences between \molora, our proposed \structone~and \structtwo.
In all of these, the base LLMs retain a frozen weight configuration, denoted as $W_0$, corresponding to the fixed linear layers within the architecture.

\par
\textbf{\molora}. 
The \molora~method serves as the foundation of our \struct.
As shown in Figure~\ref{fig:comparison}(a), the \molora~consists of a router $\theta_R^{M}$ and a set of LoRA experts $E_1, E_2, \ldots, E_n$. 
Each expert $E_i$ includes two key components: $B_i^{M}$ and $A_i^{M}$.
The dynamics of the \molora~method can be summarized by the following equations:
\begin{equation}
    \small
    s_i^{M} = {\theta_R^{M}(x_m)}_i = {softmax(W_R^{M}x_m)}_i
    \label{eq:molora-1}
\end{equation}
\begin{equation}
    \small
    y_m^{M} = E^{M}(x_m) + W_0x_m
    \label{eq:molora-2}
\end{equation}
\begin{equation}
    \small
    E^{M}(x_m) = \sum_{i=1}^{n}{s_i^{M}B_i^{M}A_i^{M}x_m} \label{eq:molora-3}
\end{equation}
In these equations, $x_m$ represents the hidden vector of the $m$-th token in the input sequence, $s_i^{M}$ denotes the routing coefficient for expert $E_i$, $W_R^{M}$ is the weight matrix of the router, and $E^{M}(\cdot)$ expresses the collective function of the experts in the \molora~module.

\par
\textbf{\struct}. 
Based on \molora, we propose a three-stage training paradigm called \struct, as illustrated in Figure~\ref{fig:training-stages}. 
In the first stage, only the universal expert is trained, while the domain-specific experts and router are deactivated. 
In the second stage, the domain-specific experts are trained individually for each corresponding task, while the parameters of the universal expert are kept frozen.
In the third stage, all the experts' parameters are fixed, and only the router is trained. 
With the \struct~paradigm, we propose two methods: \structone~and \structtwo.

\par
\textbf{\structone}. 
\structone~maintains the same architecture as \molora, as illustrated in Figure~\ref{fig:comparison}(b).
However, it implements the \struct~paradigm to separate the experts in \molora~into universal expert and domain-specific experts. 
The specific training details are as follows.
In the first stage, the universal expert $E^{flan}_*$ is trained on universal datasets. In the second stage, the domain-specific experts $E^{flan}_1, E^{flan}_2, \ldots, E^{flan}_n$ are trained on their respective domain-specific datasets. 
The forward process in this stage is formally articulated through Equations (\ref{eq:mm-1}) and (\ref{eq:mm-5}).
\begin{equation}
\small
    y_m^{flan} = E^{flan}_i(x_m) + W_0x_m
    \label{eq:mm-1}
\end{equation}
where $i\in \{1,2,\ldots,n\}$. In the third stage, the parameters of all experts are kept frozen, and only the router $\theta_R^{flan}$ is trained. 
The calculation involved in this routing determination is formally illuminated through the following equations:
\begin{equation}
\small
    s_i^{flan} = {\theta_R^{flan}(x_m)}_i = {softmax(W_R^{flan}x_m)}_i
    \label{eq:mm-2}
\end{equation}
\begin{equation}
\small
    y_m^{flan} = E^{flan}(x_m) + W_0x_m
    \label{eq:mm-3}
\end{equation}
\begin{equation}
\small
    E^{flan}(x_m) = \sum_{i}{s_i^{flan}E_i^{flan}(x_m)}
    \label{eq:mm-4}
\end{equation}
\begin{equation}
\small
    E_i^{flan}(x_m) = B_i^{flan}A_i^{flan}x_m
    \label{eq:mm-5}
\end{equation}

\par
\textbf{\structtwo}.
In order to further improve the general ability of the model, we propose \structtwo, a more advanced method that leverages the strengths of both universal and domain-specific experts. 
The architecture of \structtwo~is shown in Figure~\ref{fig:comparison}(c). 
\structtwo~integrates both the universal expert $E^{res}_*$ and the domain-specific experts $E^{res}_1, E^{res}_2, \ldots, E^{res}_n$, tuned in a balanced way to cater to both general and domain-specific tasks.
\structtwo~introduces a residual connection that allows the model to incorporate the output of universal expert directly into the final result, ensuring that critical information is preserved and enhancing model robustness.

\par
The forward process in \structtwo~module involves two stages. 
Initially, a hidden vector $h_m$ is computed using the universal expert:

\begin{equation}
\small
    h_m = B^{res}_*A^{res}_*x_m
    \label{eq:benlora-1}
\end{equation}
{where $x_m$ is the hidden vector for the $m$-th token, and $B^{res}_*$ and $A^{res}_*$ correspond to the universal expert matrices. Subsequently, the hidden vector $h_m$ is refined by the domain-specific experts with residual connection to produce the final output $y_m^{res}$:}
\begin{equation}
\small
    y_m^{res} = E^{res}(h_m) + W_0x_m + h_m
    \label{eq:benlora-3}
\end{equation}
{where the function $E^{res}(\cdot)$ represents the operation of the domain-specific experts:}
\begin{equation}
\small
    E^{res}(h_m) = \sum_{i=1}^{n}{s_i^{res}B_i^{res}LeakyReLU(A_i^{res}h_m)}
    \label{eq:benlora-4}
\end{equation}
{$s_i^{res}$ is the weight for each expert, computed as:}
\begin{equation}
\small
    s_i^{res} = {\theta_R^{res}(x_m)}_i = {softmax(W_R^{res}x_m)}_i
    \label{eq:benlora-2}
\end{equation}
This integration of a three-stage training paradigm and residual connection ensures that the \structtwo~module effectively generalizes and specializes simultaneously, thereby enhancing performance across both broad and focused applications.

\section{Experiments}

\subsection{Expert Configurations}
\par
{
A detailed comparison is conducted among the standard LoRA~\citep{hu2021lora}, \molora~\citep{zadouri2024molora}, and our newly proposed \structone~and \structtwo.
The base models selected for this study include LLaMA-2~\citep{touvron2023llama2}, Qwen~\citep{bai2023qwen}, and Yi~\citep{ai2024yi}. In the training of \struct, a batch size of 128 is utilized, encompassing 1 epoch with a learning rate of 2e-4. 
The maximum input sequence length is defined as 4096 tokens for both LLaMA-2 and Yi. 
In contrast, Qwen series has 8192 tokens due to variations in maximum positional embeddings among different model zoos. 
The intrinsic rank is configured to 16 for universal and 8 for domain-specific experts.
For the multitask results, the checkpoint selection is based on the average metrics across all tasks.
To enhance fine-tuning efficiency, we leverage libraries like HuggingFace's Transformers~\citep{wolf-etal-2020-transformers} and PEFT~\citep{peft}, based on which we design \struct.
}

\begin{table}[!htbp]
\scriptsize 
\centering
\renewcommand{\arraystretch}{1.1}
\begin{tabular*}{\linewidth}{@{\extracolsep{\fill}}lcc}
    \toprule[1.5pt]
    \textbf{Benchmark} & \textbf{Few-Shot} & \textbf{Metric} \\
    \midrule[1pt]
    GSM8K & 5 & acc \\
    Arithmetic & 0 & acc \\
    MathQA & 5 & acc \\
    HumanEval & 0 & pass@1 \\
    MBPP & 0 & pass@1 \\
    Medical & 5 & acc \\
    MedQA & 0 & acc \\
    \midrule[1pt]
    MMLU & 5 & acc \\
    C-Eval & 5 & acc \\
    \midrule[1pt]
    FinGPT-headline & 0 & acc \\
    Title-Optimization & 0 & GPT-4 Judge \\
    Keyword-Recommendation & 0 & GPT-4 Judge \\
    \bottomrule[1.5pt]
\end{tabular*}
\caption{Few-shot example numbers and evaluation metrics for benchmarks.}
\label{tab:metrics}
\end{table}

\subsection{Training Datasets}
\par
To equip our \structone~and \structtwo~with comprehensive capabilities across universal, mathematical, coding, and medical domains, the datasets \textbf{airoboros-3.2}~\footnote{https://huggingface.co/datasets/jondurbin/airoboros-3.2}, \textbf{orca-math-word-problems-200k}~\footnote{https://huggingface.co/datasets/microsoft/orca-math-word-problems-200k}, \textbf{CodeAlpaca-20k}~\footnote{https://huggingface.co/datasets/sahil2801/CodeAlpaca-20k}, and \textbf{MedQA}~\citep{jin2019pubmedqa} are integrated. 

\par
In order to evaluate the pluggability of our methods, we fine-tune the baselines, \structone, and \structtwo~on three datasets from different domains: \textbf{FinGPT-headline}~\footnote{https://huggingface.co/datasets/FinGPT/fingpt-headline} from the finance domain, and \textbf{Title-Optimization} and \textbf{Keyword-Recommendation} from the e-commerce domain. The Title-Optimization and Keyword-Recommendation datasets are sourced from real-world requirements on alibaba.com~\footnote{https://www.alibaba.com/}, a leading e-commerce platform. By fine-tuning on these diverse datasets, we aim to demonstrate the adaptability and effectiveness of \structtwo~in various domain-specific applications, showcasing its modular design and ability to capture both general and domain-specific knowledge.

\begin{table}[!htbp]
\centering
\scriptsize 
\renewcommand{\arraystretch}{0.9} 
\begin{adjustbox}{max width=\linewidth}
\begin{tabular*}{\linewidth}{@{\extracolsep{\fill}}cccc}
    \toprule[1.5pt]
    \textbf{Base Model} & \textbf{Method} & \textbf{MMLU} & \textbf{C-Eval} \\
    \midrule[1pt]
    \multirow{4}{*}{Qwen-7B} & Not fine-tuned & \textbf{58.21} &	62.1 \\
    ~ & \molora~&  55.77 &	61.44 \\
    \cmidrule(lr){2-4}
    ~ & \structone & 56.16 & 62.29 \\
    ~ & \structtwo & 57.65 & \textbf{62.34} \\
    \midrule[1pt]
    
    \multirow{4}{*}{LLaMA-2-7B} & Not fine-tuned & 45.91 &	34.02 \\
    ~ & \molora~& 	47.45 &	35.95  \\
    \cmidrule(lr){2-4}
    ~ & \structone & 45.65 & 34.22 \\
    ~ & \structtwo & \textbf{48.23} &	\textbf{36.18} \\
    \midrule[1pt]
    
    \multirow{4}{*}{Yi-6B} & Not fine-tuned & 63.30 &	73.63 \\
    ~ & \molora~&  63.11 &	73.17 \\
    \cmidrule(lr){2-4}
    ~ & \structone & 62.17 & 72.33 \\
    ~ & \structtwo & \textbf{63.41} &	\textbf{74.15} \\
    \midrule[1pt]

    \multirow{4}{*}{Qwen-14B} & Not fine-tuned & 66.89 & \textbf{70.87} \\
    ~ & \molora~&  \textbf{67.21} &	70.35 \\
    \cmidrule(lr){2-4}
    ~ & \structone & 65.98 & 69.82 \\
    ~ & \structtwo &  66.58 &	70.13 \\
    \midrule[1pt]

    \multirow{4}{*}{LLaMA-2-13B} & Not fine-tuned & 54.92 &	38.11 \\
    ~ & \molora~& 56.08 &	40.34 \\
    \cmidrule(lr){2-4}
    ~ & \structone & \textbf{57.01} & 39.22 \\
    ~ & \structtwo & 56.23 & \textbf{40.94} \\
    \midrule[1pt]

    \multirow{4}{*}{Yi-9B} & Not fine-tuned & 68.10 &	\textbf{70.57} \\
    ~ & \molora~&  67.70 &	69.83\\
    \cmidrule(lr){2-4}
    ~ & \structone & 66.07 & 68.41\\
    ~ & \structtwo &  \textbf{68.13} &	69.83 \\
    \bottomrule[1.5pt]
\end{tabular*}
\end{adjustbox}
\caption{Experimental results of different methods on universal benchmarks.}
\label{tab:task-unverisal}
\end{table}

\begin{table*}[htbp]
\centering
\scriptsize 
\renewcommand{\arraystretch}{1.1} 
\begin{adjustbox}{max width=\textwidth}
\begin{tabular*}{\textwidth}{@{\extracolsep{\fill}}cccccccccc|c}
    \toprule[1.5pt]
    \textbf{Base Model} & \textbf{Method} & \textbf{Avg.}  & \textbf{GSM8K} & \textbf{Arithmetic} & \textbf{MathQA} & \textbf{HumanEval} & \textbf{MBPP} & \textbf{Medical} & \textbf{MedQA} & \makecell[c]{\textbf{FinGPT} \\ \textbf{headline}} \\
    \midrule[1pt]
    \multirow{4}{*}{Qwen-7B} & Not fine-tuned & 44.65 &	46.63 &	56.65 &	35.48 &	21.95 &	32.00 &	76.00 &	43.83 &	74.91 \\
    ~ & \molora~& 49.92 & 47.38 & 84.05 & 36.88 & 22.56 & 32.00 & \textbf{80.20} & \textbf{46.34} & 75.41 \\
    \cmidrule(lr){2-11}
    ~ & \structone~& 50.66 &	\textbf{48.36} &	88.12 &	36.41 &	26.22 &	\textbf{32.60} &	79.00 &	43.93 & 76.61 \\
    ~ & \structtwo & \textbf{50.85} &	45.87 &	\textbf{89.37} &	\textbf{37.99} &	\textbf{28.05} &	31.02 &	79.60 &	44.06 & \textbf{80.61} \\
    \midrule[1pt]
    
    \multirow{4}{*}{LLaMA-2-7B} & Not fine-tuned & 27.45 &	13.72 &	6.89 &	29.41 &	14.63 &	18.00 &	77.60 &	31.89 & 22.39 \\
    ~ & \molora~& 37.05 &	16.75 &	57.20 &	30.51 &	\textbf{16.46} &	20.40 &	\textbf{82.20} &	\textbf{35.82} & 32.38 \\
    \cmidrule(lr){2-11}
    ~ & \structone~& 37.37 &	17.43 &	59.38 &	30.61 &	15.85 &	23.20 &	81.80 &	33.30 & 24.89 \\
    ~ & \structtwo & \textbf{37.86} &	\textbf{21.61} &	\textbf{67.59} &	\textbf{31.26} &	12.80 &	\textbf{24.00} &	81.20 &	26.56 & \textbf{33.83} \\
    \midrule[1pt]
    
    \multirow{4}{*}{Yi-6B} & Not fine-tuned & 38.04 &	33.81 &	39.92 &	35.41 &	14.63 &	23.00 &	70.00 &	49.49 & 64.92 \\
    ~ & \molora~& 48.58 &	\textbf{36.42} &	93.50 &	\textbf{36.78} &	16.46 &	23.80 &	81.20 &	\textbf{51.92} & 65.96 \\
    \cmidrule(lr){2-11}
    ~ & \structone~& 47.98 &	35.17 &	92.57 &	36.18 &	16.46 &	\textbf{24.80} &	80.20 &	50.50 & 61.77 \\
    ~ & \structtwo & \textbf{48.70} &	34.57 &	\textbf{93.63} &	36.15 &	\textbf{17.07} &	23.40 &	\textbf{85.60} &	50.51 & \textbf{73.26} \\
    \midrule[1pt]
\end{tabular*}
\end{adjustbox}
\caption{Experimental results of \molora, \structone, and \structtwo~on domain-specific and FinGPT-headline (finance) benchmarks.}
\label{tab:pluggability-domain-specific}
\end{table*}

\begin{table}[!htbp]
\centering
\scriptsize 
\renewcommand{\arraystretch}{1.1} 
\begin{adjustbox}{max width=0.5\textwidth}
\begin{tabular}{cc|cc|cc}
    \toprule[1.5pt]
    \textbf{Base Model} & \textbf{Method} & \textbf{Avg.} & \textbf{T.O.} & \textbf{Avg.} & \textbf{K.R.} \\
    \midrule[1pt]
    \multirow{4}{*}{Qwen-7B} & Not fine-tuned & 44.65 & 6.23 & 44.65 & 5.28\\
    ~ & \molora~& 49.89 & 6.94 & 48.64 & 5.92 \\
    \cmidrule(lr){2-6}
    ~ & \structone~& 50.19 & 5.44 & 49.59 & 6.78\\
    ~ & \structtwo & \textbf{51.17} & \textbf{7.28} & \textbf{50.29} & \textbf{7.02} \\
    \midrule[1pt]
    
    \multirow{4}{*}{LLaMA-2-7B} & Not fine-tuned & 27.45 & 2.76 & 27.45 & 4.25 \\
    ~ & \molora~& 35.35 & 3.54 & 36.23 & 5.98\\
    \cmidrule(lr){2-6}
    ~ & \structone~& 37.21 & 6.48 & 37.33 & 6.52\\
    ~ & \structtwo & \textbf{38.80} & \textbf{6.62} & \textbf{38.12} & \textbf{7.37}\\
    \midrule[1pt]
    
    \multirow{4}{*}{Yi-6B} & Not fine-tuned & 38.04 & 3.01 & 38.04 & 5.45\\
    ~ & \molora~& 45.91 & 3.92 & 44.37 & 5.78\\
    \cmidrule(lr){2-6}
    ~ & \structone~& 47.93  &5.92 & 46.59  & 6.38\\
    ~ & \structtwo & \textbf{48.28}& \textbf{6.94} & \textbf{47.88}& \textbf{7.58}\\
    \midrule[1pt]
\end{tabular}
\end{adjustbox}
\caption{Experimental results of methods on T.O. and K.R. (e-commerce) benchmarks. Avg. denotes the average performance of different methods on domain-specific benchmarks. T.O. denotes the Title Optimization task and K.R. the Keyword Recommendation task.}
\label{tab:domain-specific}
\end{table}

\subsection{Evaluation Benchmarks and Metrics}
\par
To comprehensively assess the performance of various methods, we conduct evaluations across a diverse set of benchmarks. 
Domain-specific performance is evaluated by testing mathematical abilities on \textbf{GSM8K}~\citep{cobbe2021gsm8k}, \textbf{Arithmetic}~\citep{brown2020gpt}, and \textbf{MathQA}~\citep{amini2019mathqa}, coding skills on \textbf{HumanEval}~\citep{chen2021codexhumaneval} and \textbf{MBPP}~\citep{austin2021mbpp}, and medical knowledge on \textbf{MedQA}~\citep{jin2020disease} and the \textbf{Medical}~\citep{jin2019pubmedqa} dataset. 
General capabilities are measured via \textbf{MMLU}~\citep{hendrycks2021mmlu} and \textbf{C-Eval}~\citep{huang2023ceval} benchmarks, which both cover a wide range of tasks.
To evaluate the pluggability and adaptability of different methods on new domain-specific tasks, we test their performance on the \textbf{FinGPT-headline}~\citep{yang2023fingpt} dataset from the finance domain, as well as the \textbf{Title-Optimization} and \textbf{Keyword-Recommendation} datasets from the e-commerce domain. 

\par
Title optimization and keyword recommendation are critical tasks in e-commerce that aim to enhance product visibility and market responsiveness. 
These tasks involve integrating high-exposure queries from specific leaf categories into product titles to refine original titles and generate new keywords, ultimately achieving a higher Click-Through Rate (CTR).
By evaluating the methods of these real-world e-commerce tasks, we can assess their effectiveness in capturing domain-specific knowledge and potential for practical application in industry settings.
The specific evaluation metrics used for each benchmark are summarized in Table~\ref{tab:metrics}, providing a clear overview of the performance measures employed in our experiments.

\subsection{Main Experimental Results}
\par
Our experimental results yield several significant observations that demonstrate the robustness and effectiveness of the proposed approach, providing valuable insights into its performance across various benchmarks and real-world applications.

\par
\textbf{Superior Advancement over Baselines}:
Table~\ref{tab:main-table-domain-specific} highlights the significant performance improvements achieved by our proposed paradigm across Qwen, LLaMA-2, and Yi.
Models that are fine-tuned with our paradigm outperform the base models by an average of 16.6\% and surpass the performance of \molora~by 6.3\% on average.
Notably, Yi demonstrates the most substantial improvement, with an impressive average increase of 10.9\% over \molora. 

\par
Further analysis reveals that performance advancements are more pronounced in smaller-scale models than in their larger counterparts, e.g., 4.9\% for Qwen-7B while 2.9\% for Qwen-14B. This indicates that small-scale models with fewer parameters and inadequate training are more prone to losing general capability when learning multiple tasks, while residual connections can effectively mitigate this problem.


\par


Moreover, \structone~does not consistently outperform \molora, suggesting that it has the issue of decreased general capabilities (for example, the arithmetic benchmark of LLaMA-2-13B dropped sharply due to the decline in text understanding ability). In contrast, \structtwo~addresses this issue by introducing residual connections for general and expert modules, leading to more stable performance and significant improvements over \molora~and \structone.


\par
Despite \structtwo~demonstrates overall strong performance, it faces challenges with GSM8K and MedQA tasks, likely due to the mismatch between pre-training data and task-specific requirements. We recognize these limitations and leave them for further research.


\par
\textbf{Excellent Robustness on Comprehensive Benchmarks}:
In order to determine whether the general capability of \structtwo~trained on multiple tasks will decline, we conduct experiments using the base, \molora, and the \structtwo~model on the comprehensive benchmarks MMLU and C-Eval.

\par
The results in Table~\ref{tab:task-unverisal} indicate that the average performance of \structtwo~across multiple models is about 1\% higher than that of \molora~and the base model, suggesting that the model's general capability is maintained and even partially improved through residual connection.

\par
\textbf{Flexible Pluggability over Baselines}:
To showcase \structtwo's pluggability, we introduce the finance domain (FinGPT-headline) in addition to the initial domains of mathematics, coding, and medical care. Then, we retrained \molora, \structone, and \structtwo, respectively. \molora~is trained from scratch on the combined dataset, while \structone~and \structtwo~only require training a new financial expert and the router. This results in \structone~and \structtwo~using only 19.8\% and 37.3\% of the training parameters and data compared to \molora, respectively.

\par
The results in Table~\ref{tab:pluggability-domain-specific} indicate that \structtwo~achieves the best average multi-task performance among the three models, with an average improvement of 8.0\% in the financial task. 
Notably, the overall improvement of Yi-6B is more significant, exceeding 11.0\%, due to the fewer parameters and relatively balanced pre-training data. 
\molora~encounters issues with data balance, requiring numerous experiments to adjust the data ratio for each task to achieve the best overall performance when new domain-specific tasks are introduced, which is time-consuming and labor-intensive.

\begin{figure*}[!htbp]
  \includegraphics[width=0.48\linewidth]{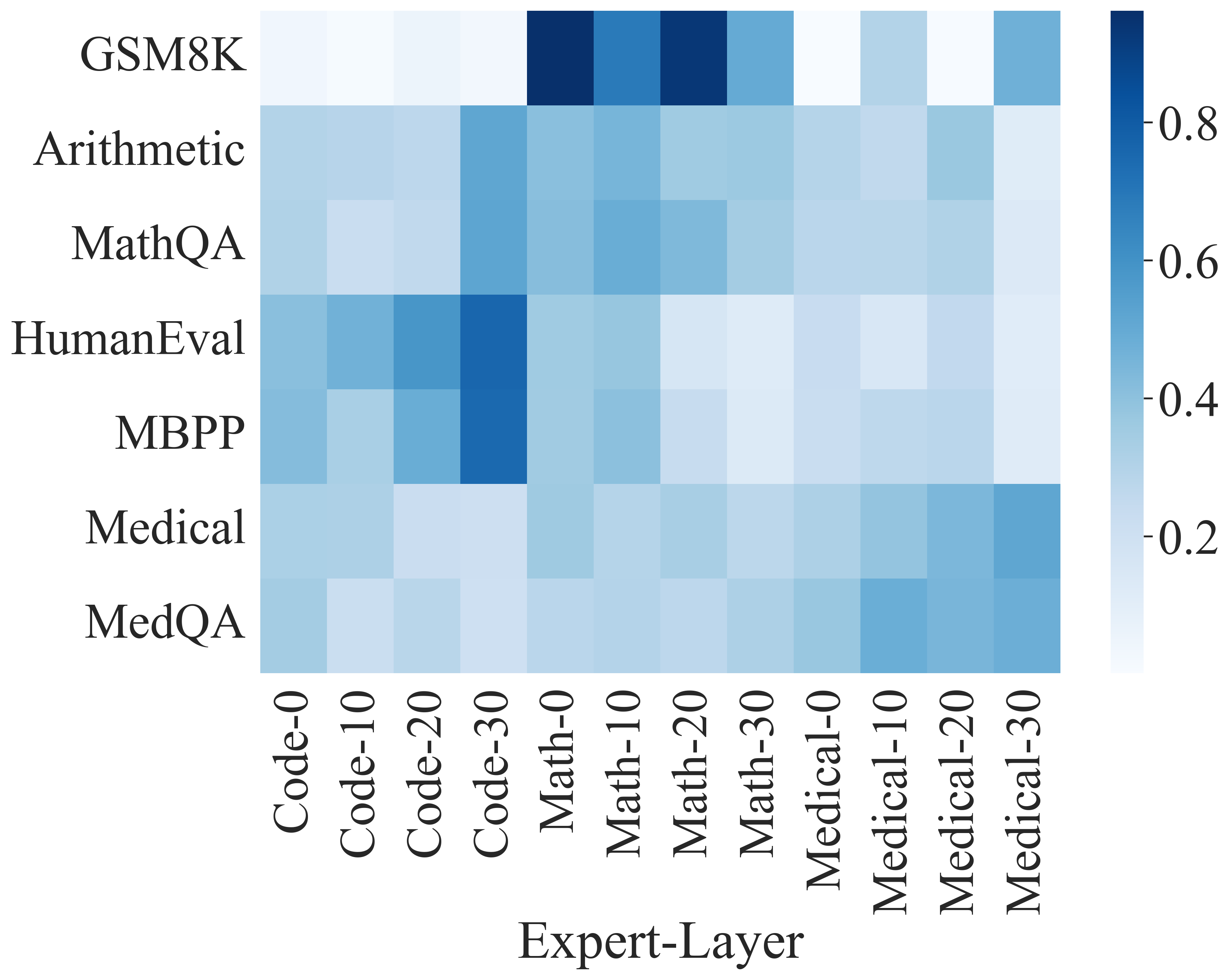} \hfill
  \includegraphics[width=0.48\linewidth]{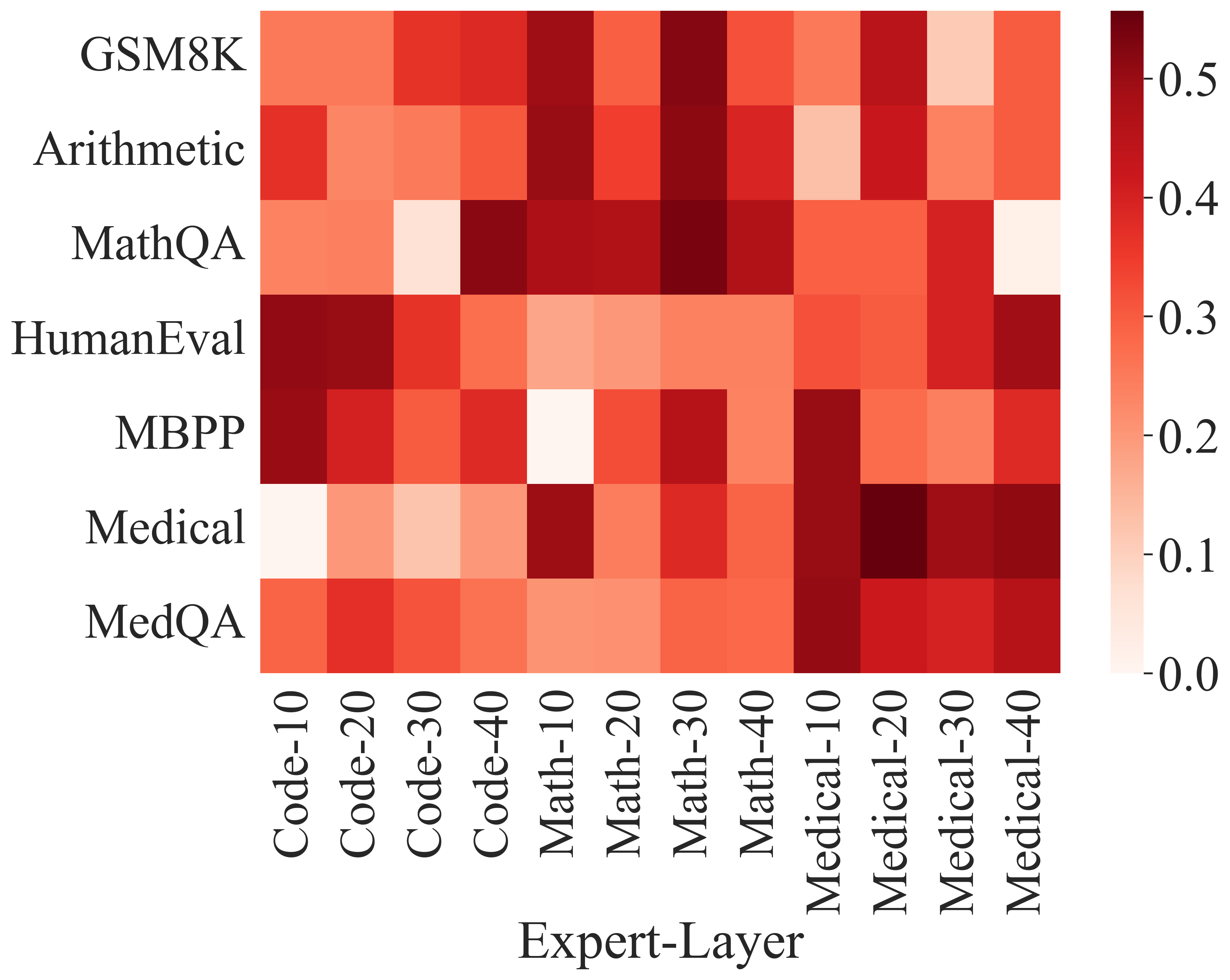}
  \caption {Router distributions of \structtwo~based on Yi-6B (left) and Qwen-14B (right) on domain-specific tasks.}
    \label{fig:router-out}
\end{figure*}

\par
\textbf{Outstanding Performance in E-Commerce}:
To assess \struct's practical applicability in e-commerce, we introduce title optimization and keyword recommendation tasks, which involve refining titles and generating keywords using high-exposure queries to enhance readability and include more key points. 
We employ GPT-4 to evaluate the optimized titles and keywords across five dimensions: helpfulness, relevance, accuracy, readability, and fluency. 
Each dimension is scored 0, 1, or 2, with a maximum total score of 10.

\par
Table 5 demonstrates that \structtwo~significantly improves performance on title optimization and keyword recommendation benchmarks, with gains of 44.7\% and 24.3\% over \molora, respectively. Moreover, \structtwo~maintains superior performance on the original multi-task benchmarks. These results highlight \structtwo's potential for e-commerce applications and adaptability to new tasks under resource constraints.

\subsection{Analysis on Domain-specific Experts Allocation}
\par
To further analyze \structtwo, router distributions for domain-specific experts based on Yi-6B and Qwen-14B are visualized in Figure~\ref{fig:router-out}. 
Models in Table~\ref{tab:main-table-domain-specific} are reused, and we select layer 0-10-20-30 and 10-20-30-40 for Yi-6B and Qwen-14B, respectively. 

\par
The results indicate that for both the Yi and Qwen models, the router within the \struct~paradigm allows various experts to concentrate on their own domain. However, the interpretation of expert assignments varies across different layers in different models due to the model's training data and method. 
For instance, Yi's deeper layers focus more on separating experts, while Qwen in the shallower layers.

\section{Conclusion}
\par
In this paper, we introduce \struct, a novel multi-stage training PEFT MoE paradigm that enhances efficiency and domain-specific adaptation for LLMs. By integrating universal and domain-specific experts through a three-stage training methodology, \struct~optimizes both generalization and specialized performance. Experiments on various open-source LLMs, such as LLaMA-2, Qwen, and Yi, demonstrate that \struct~outperforms existing methods, achieving over 80\% reduction in training costs and a 5\% performance improvement. These results highlight \struct's potential as a scalable and efficient solution for fine-tuning LLMs, paving the way for future advancements in NLP.

\section*{Acknowledgement}
This work was supported in part by the National Natural Science Foundation of China under Grants U20B2065, U22B2036, 62372380, and 62103374, National Key Research and Development Project under Grant 2022YFB3104005, and the Natural Science Basic Research Program of Shaanxi (Program No.2024JC-YBMS-513), and Key Research and Development Program of Zhejiang Province under Grants 2024C01025.

\section*{Limitations}
\par
While our proposed \struct~paradigm shows significant advancements in parameter efficiency and multi-task adaptability for LLMs, there are still some limitations that need to be addressed. 
Despite the overall strong performance of \structtwo, it shows sub-optimal results on certain benchmarks like GSM8K and MedQA. 
This may be due to discrepancies between the model's pre-training data and the specific task datasets, requiring further investigation to identify the root causes and develop targeted solutions. 
Our experiments also focus on a limited set of language models (LLaMA-2, Qwen, Yi) and domain-specific tasks (mathematics, coding, medical, finance, e-commerce). 
To establish stronger generalizability, it would be valuable to extend our evaluations to a broader range of base models and diverse task domains.
Furthermore, the current study primarily emphasizes the pluggability and training efficiency of \struct~when incorporating new domain experts. 
However, the scalability and robustness of this approach when integrating a larger number of experts require further exploration and stress testing.

\par
Future research directions include investigating techniques to mitigate performance degradation on specific benchmarks, conducting comprehensive evaluations on a wider range of models and tasks, exploring the scalability limits of expert integration, streamlining the multi-stage training process, and enhancing the interpretability of the router's decision-making. 
By acknowledging these limitations and outlining potential avenues for future work, we aim to provide a balanced perspective on the current state of our research and highlight opportunities for further advancements in PEFT for LLMs.


\newpage

\begin{table*}[!htbp]
\centering
\scriptsize 
\renewcommand{\arraystretch}{0.9} 
\begin{adjustbox}{max width=\textwidth}
\begin{tabular*}{\textwidth}{@{\extracolsep{\fill}}cccccccccc}
    \toprule[1.5pt]
    \textbf{Base Model} & \textbf{Method} & \textbf{Avg.}  & \textbf{GSM8K} & \textbf{Arithmetic} & \textbf{MathQA} & \textbf{HumanEval} & \textbf{MBPP} & \textbf{Medical} & \textbf{MedQA}\\
    \midrule[1pt]
    \multirow{2}{*}{Qwen-7B} & \structtwo~& \textbf{51.36} &	\textbf{46.63} &	\textbf{90.37} &	\textbf{37.98} &	\textbf{25.00} &	33.00 &	\textbf{82.00} &	\textbf{44.55} \\
    ~ & {\struct~w/o. Res} & 49.65 &	46.47 &	85.35 &	35.24 &	24.39 &	\textbf{33.60} &	78.20 &	44.31 \\
    \midrule[1pt]
    
    \multirow{2}{*}{LLaMA-2-7B} & \structtwo & \textbf{39.62} &	\textbf{22.37} &	\textbf{70.66} &	\textbf{31.73} &	15.24 &	\textbf{22.80} &	\textbf{85.20} &	29.31 \\
    ~ & {\struct~w/o. Res} & 37.85 &	20.40 &	61.82 &	30.45 &	\textbf{16.46} &	22.80 &	79.00 &	\textbf{34.01} \\
    \midrule[1pt]
    
    \multirow{2}{*}{Yi-6B} & \structtwo & \textbf{48.61} &	34.50 &	\textbf{92.72} &	\textbf{36.29} &	\textbf{16.46} &	\textbf{24.40} &	\textbf{85.80} &	50.12  \\
    ~ & {\struct~w/o. Res} & 45.60 &	\textbf{34.87} &	90.01 &	35.94 &	14.24 &	15.80 &	77.60 &	\textbf{50.74} \\
    \bottomrule[1.5pt]
\end{tabular*}
\end{adjustbox}
\caption{Experimental results of \structtwo~and \struct~w/o. Res on domain-specific benchmarks.}
\label{tab:domain-specific-without-res}
\end{table*}

\appendix
\section{Analysis on the Residual Connection}
\label{appendix: res-con}

\par
The results in Table~\ref{tab:domain-specific-without-res} validate the importance of the residual connection in the \structtwo~method. Comparing \structtwo~with its non-residual counterpart reveals the residual connection's role in enhancing domain-specific tasks while preserving general language understanding. 

\par
The residual connection's impact varies among models. For instance, Qwen-7B and Yi-6B models show significant score improvements of 1.71 and 3.01 points, respectively, whereas LLaMA-2-7B shows a smaller gain of 1.77 points. This suggests that the benefits may be model-specific, meriting further investigation.

\par
In domain-specific tasks, \structtwo~excels, particularly in mathematics and medical fields. For example, in Arithmetic and Medical datasets, \structtwo~exceeds its non-residual variant by over 5 points, signifying the residual connection's role in effective knowledge transfer.

\par
However, in some tasks like MBPP and MedQA, the non-residual model slightly outperforms \structtwo. This nuance suggests a need to further analyze the residual connection's mechanism across various tasks to improve the model's robustness.

\par
In conclusion, the findings affirm the \structtwo~method's efficacy. Residual connections significantly enhance overall performance on domain-specific tasks, offering a promising avenue for future enhancements in the PEFT paradigm. 
Continued exploration of residual connections in multi-task learning is expected to yield more powerful and versatile language models.

\section{GPT-4 Judge Prompt for E-commerce Tasks}
\label{appendix: gpt4-judge-pt}

\begin{figure}[t]
  \includegraphics[width=\columnwidth]{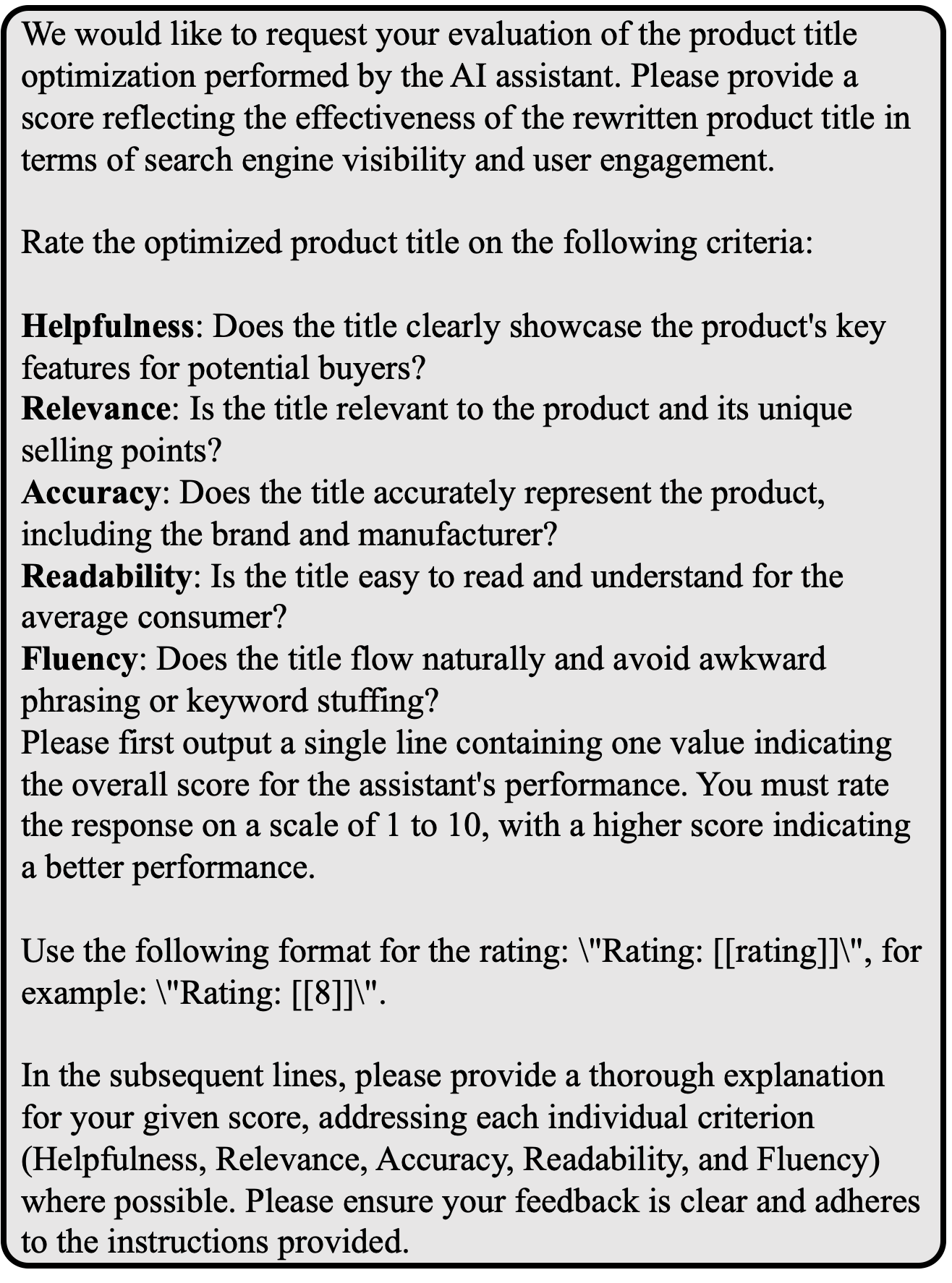}
  \caption{The GPT-4 judge prompt for Title-Optimization task.}
  \label{fig:titleoptimization}
\end{figure}

\begin{figure}[t]
  \includegraphics[width=\columnwidth]{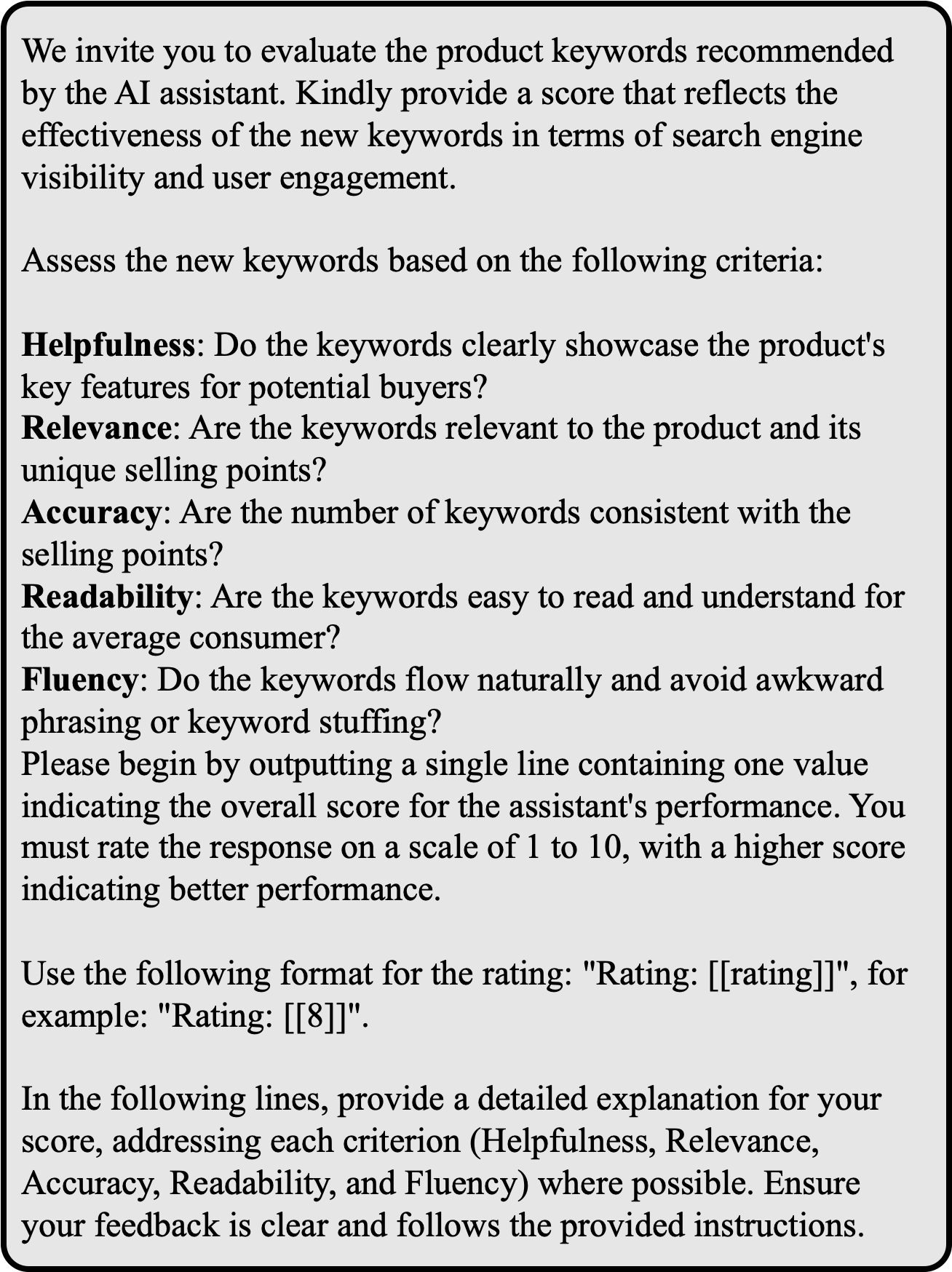}
  \caption{The GPT-4 judge prompt for Keyword-Recommendation task.}
  \label{fig:keyword}
\end{figure}

\end{document}